\documentclass[11pt,a4paper]{article}
\usepackage[margin=1in]{geometry}
\usepackage[T1]{fontenc}
\usepackage{lmodern,amsmath,amssymb,graphicx,booktabs,array,placeins,flafter}
\DeclareFontFamily{T1}{gtmathbold}{}
\DeclareFontShape{T1}{gtmathbold}{m}{n}{<-> ec-lmbx10}{}
\DeclareMathAlphabet{\mathbf}{T1}{gtmathbold}{m}{n}
\usepackage[font=small,labelfont=bf,justification=raggedright,singlelinecheck=false]{caption}
\usepackage{xurl}
\usepackage[hidelinks]{hyperref}
\usepackage{microtype}
\usepackage{authblk}
\title{GateTrim: Adaptive-Rank Quantization for Large Language Models}
\author{Hongwei Ren}
\affil{renhongwei2002@qq.com}
\author{Jincheng Ren}
\affil{Jincheng.ren.gm@gmail.com}
\author{Yuchuan Zhou}
\affil{zhouyuchuancd@qq.com}
\author{Jingchun Ling\thanks{Corresponding author: Jingchun Ling (lingjingchun1990@qq.com).}}
\affil{lingjingchun1990@qq.com}
\date{29 October 2026}
\hypersetup{pdftitle={GateTrim: Adaptive-Rank Quantization for Large Language Models},pdfauthor={Hongwei Ren, Jincheng Ren, Yuchuan Zhou, Jingchun Ling}}
\begin{document}\maketitle
\begin{abstract}We introduce GateTrim, an adaptive-rank post-training quantization method for large language models that learns matrix-specific ranks through gate optimization and threshold-based rank selection. GateTrim standardizes the low-rank factors of a quantized low-rank--residual representation, learns a separate gate vector for each target weight matrix, and uses a shared threshold to automatically determine its retained rank. The resulting ranks can differ across matrices, and the compact quantized representations are subsequently reconstructed. We evaluate GateTrim under W4A8 on Qwen3-4B/8B/14B, Llama-3.2-3B-Instruct, and Llama-3.1-8B-Instruct. Across all five models, we observe a counterintuitive finding: increasing the threshold from $\tau=0.3$ to $\tau=0.7$ reduces the mean retained rank by 46.60\%--67.89\% while simultaneously lowering WikiText-2 perplexity by 0.59\%--5.17\%. Threshold 0.7 achieves the lowest mean retained rank and final perplexity among the three tested thresholds, 0.3, 0.5, and 0.7, and reduces measured target-export size by 1.50\%--4.25\% relative to 0.3. GateTrim uses an $8\times512$-token fitting pool and a separate development pool of the same size. Its Qwen3-4B/8B/14B perplexities at $\tau=0.7$ are numerically below TwinQuant's published W4A8 values, reported with a $128\times2048$-token calibration pool, in a cross-study comparison. On Qwen3-8B, the threshold-0.7 configuration already has the lowest perplexity before reconstruction under both tested initialization seeds; reconstruction improves it further without increasing retained rank or export size.\end{abstract}

\FloatBarrier\section{Introduction}

Language model quantization reduces the precision used to represent weights and activations. Its effectiveness depends on how quantization error propagates through the model, as well as on the storage and execution costs of the resulting representation. A useful design represents a weight matrix through a dense quantized residual and an additional low-rank branch. The residual preserves a full-dimensional representation, while the low-rank branch provides complementary capacity for reconstructing the original operator.

This design introduces a capacity-allocation problem. A fixed rank assigns the same number of low-rank components to every matrix, although different attention and feed-forward projections may require different amounts of additional capacity. Increasing the rank adds parameters and computation; reducing it may remove components that are useful after quantization. The appropriate allocation therefore depends on the quantized computation and its reconstruction objective, rather than solely on a decomposition performed before quantization.

Continuous gates provide a way to adjust the contribution of individual components. However, small gate values alone do not reduce stored tensor dimensions. Storage savings require explicit rank reduction and compact serialization. Compaction can also change weight-quantization groups and the groups used to quantize the intermediate activation. A compact representation is consequently not guaranteed to behave like the original representation with a subset of contributions set to zero.

GateTrim learns the retained rank of each target weight matrix through its own gate parameters. A target matrix is an individual attention or MLP projection, rather than an entire Transformer block or a quantization group. Stage 1 reconstructs a quantized low-rank--residual representation while standardizing each row of $U$ and each column of $V$. Stage 2 learns gate amplitudes with the factors and normalization statistics fixed. A shared threshold then converts the learned gates into matrix-specific retained ranks. Stage 3 reconstructs each compact quantized representation at its selected rank. This procedure learns rank through continuous gate optimization followed by discrete rank selection; the threshold is a specified hyperparameter.

Across the five models evaluated under W4A8, increasing the threshold from 0.3 to 0.7 lowers both the mean retained rank and final WikiText-2 perplexity. Threshold 0.7 gives the lowest mean retained rank and final perplexity among the three tested thresholds in every model. For Qwen3-8B, this preference is already present after compact requantization but before reconstruction (q0) under both tested initialization seeds. Reconstruction further lowers perplexity without changing retained rank or export size. The results show that retaining more low-rank components does not necessarily improve the quality of the quantized model.

Our evaluation also examines quantization with a small calibration fitting pool. GateTrim uses eight 512-token fitting sequences and eight separate development sequences, with no gradient updates on the development pool. At $\tau=0.7$, GateTrim achieves W4A8 WikiText-2 perplexities of 15.41, 11.24, and 9.49 on Qwen3-4B, Qwen3-8B, and Qwen3-14B, respectively. These are numerically below TwinQuant's published values of 18.3, 11.5, and 10.5, obtained with 128 calibration sequences of 2,048 tokens each [1]. This comparison places our results in the context of published work; the experimental settings of the two studies are detailed in Section 5.3.

The contributions are:

\begin{enumerate}

\item \textbf{Learning matrix-specific ranks for quantization.} We introduce GateTrim, an adaptive-rank post-training quantization method that combines factor standardization with learned gates. A shared threshold automatically determines each target matrix's retained rank from its learned gates, and the resulting compact quantized representation is reconstructed.

\item \textbf{Lower rank with lower perplexity.} We identify a counterintuitive finding across five model configurations: threshold 0.7 produces a lower mean retained rank and lower WikiText-2 perplexity than threshold 0.3. The preference for 0.7 is also observed before reconstruction on Qwen3-8B under both tested initialization seeds.

\end{enumerate}

\FloatBarrier\section{Background and Related Work}

\FloatBarrier\subsection{Weight and Activation Quantization}

For a group of real-valued entries $x$, symmetric quantize--dequantize computation can be expressed as

\begin{equation*}
\mathcal Q_b(x)=\widetilde\alpha\,\operatorname{clip}\left(\operatorname{round}(x/\alpha),q_{\min},q_{\max}\right),
\end{equation*}

where $\alpha$ determines the integer codes and $\widetilde\alpha$ is the stored reconstruction scale. These quantities can differ when scales are serialized at reduced precision.

SmoothQuant uses equivalent channel scaling to redistribute quantization difficulty between activations and weights. OmniQuant learns clipping and equivalent transformations through reconstruction-based calibration. FlatQuant learns invertible transformations together with scaling and clipping parameters; SpinQuant learns rotations that improve the suitability of tensors for quantization. These methods optimize the representation supplied to the quantizer. SmoothQuant~\cite{ref4}, OmniQuant~\cite{ref5}, FlatQuant~\cite{ref6}, SpinQuant~\cite{ref7}.

Our setting quantizes both factors and the residual of a decomposed weight representation, as well as the input and the intermediate activation inside the low-rank branch.

\FloatBarrier\subsection{Low-Rank--Residual Representations}

Low-rank components can represent information that is difficult to preserve at low precision. LQ-LoRA studies low-rank-plus-quantized decomposition for efficient language model finetuning, including weighted decomposition objectives. ZeroQuant-V2 investigates low-rank compensation for post-training quantization. Their objectives and numerical configurations differ from the present setting. LQ-LoRA~\cite{ref2}, ZeroQuant-V2~\cite{ref3}.

TwinQuant learns quantization-friendly decomposed subspaces through component-specific transformations and develops a fused execution kernel. GateTrim learns matrix-specific retained ranks from gates in a standardized quantized representation and reconstructs the compact factors at those ranks. A direct performance comparison requires matched checkpoints, numerical configurations, and evaluation protocols. TwinQuant~\cite{ref1}.

\FloatBarrier\subsection{Learned Gates and Retained Rank}

A continuous gate controls the amplitude of each latent component of the low-rank branch. Gate parameters are learned from calibration data. Applying a shared threshold to the learned gates determines how many components each matrix retains, so the final ranks can vary across matrices. A small gate value alone does not reduce stored tensor dimensions; compact tensors must be formed at the selected rank.

These thresholds act on gate amplitudes and differ from clipping thresholds for a quantizer's dynamic range. Learned components are not constrained to remain the original singular vectors or orthogonal components. We use retained rank to denote the retained dimension of the low-rank branch, not the algebraic rank of the complete weight representation, which also contains a dense residual. The selected rank is fixed during reconstruction and inference.

\FloatBarrier\section{Motivation}

\FloatBarrier\subsection{Retained Rank Controls Only Part of the Representation Cost}

Let $W\in\mathbb R^{m\times d}$ be represented by a dense residual $R\in\mathbb R^{m\times d}$ and factors $U\in\mathbb R^{m\times r}$ and $V\in\mathbb R^{r\times d}$. With four-bit codes for all three tensors, idealized code storage is

\begin{equation*}B_{\mathrm{codes}}(r)=\frac{md+r(m+d)}{2}\quad\text{bytes}.\end{equation*}

Reducing $r$ decreases factor storage, while the dense residual remains. The achievable total reduction depends on matrix shape and the residual's contribution. Average retained rank alone therefore cannot quantify actual compression. Both ranks and measured bytes must be reported.

\FloatBarrier\subsection{Continuous Gates Must Become Compact Quantized Tensors}

A small gate does not remove factor entries from storage. Physical compression requires selecting indices and extracting the corresponding columns and rows. This conversion can change quantization groups:

\begin{equation*}\mathcal Q_4(U_{:,I})\neq[\mathcal Q_4(U)]_{:,I}.\end{equation*}

The selected entries can be quantized differently when their group contents and scales change. The intermediate-activation quantizer can change as well. Comparing a parent representation with its compact version can therefore mix rank reduction, regrouping, and subsequent optimization. We separate these effects through explicit endpoints.

\FloatBarrier\subsection{Reconstruction Evaluates the Compact Representation}

The parameters before rank selection were calibrated for a different set of components. After selection, the surviving factors and residual can compensate for some removed contributions. We treat the immediately compacted representation as an intermediate state and evaluate it after a fixed reconstruction budget, keeping its retained indices unchanged.

This comparison determines whether reconstruction improves accuracy without adding retained capacity. It also exposes the distinction between local reconstruction quality and final model perplexity: one can improve without establishing a uniform bound on the other.

\FloatBarrier\section{Method}

GateTrim connects learned component amplitudes with physical compaction and reconstruction. Figure 1 shows the gated operator and the calibration stages; quantizers below specify the W4A8 realization used in this study.

\begin{figure}[!htbp]\centering
\includegraphics[width=\linewidth,height=0.82\textheight,keepaspectratio]{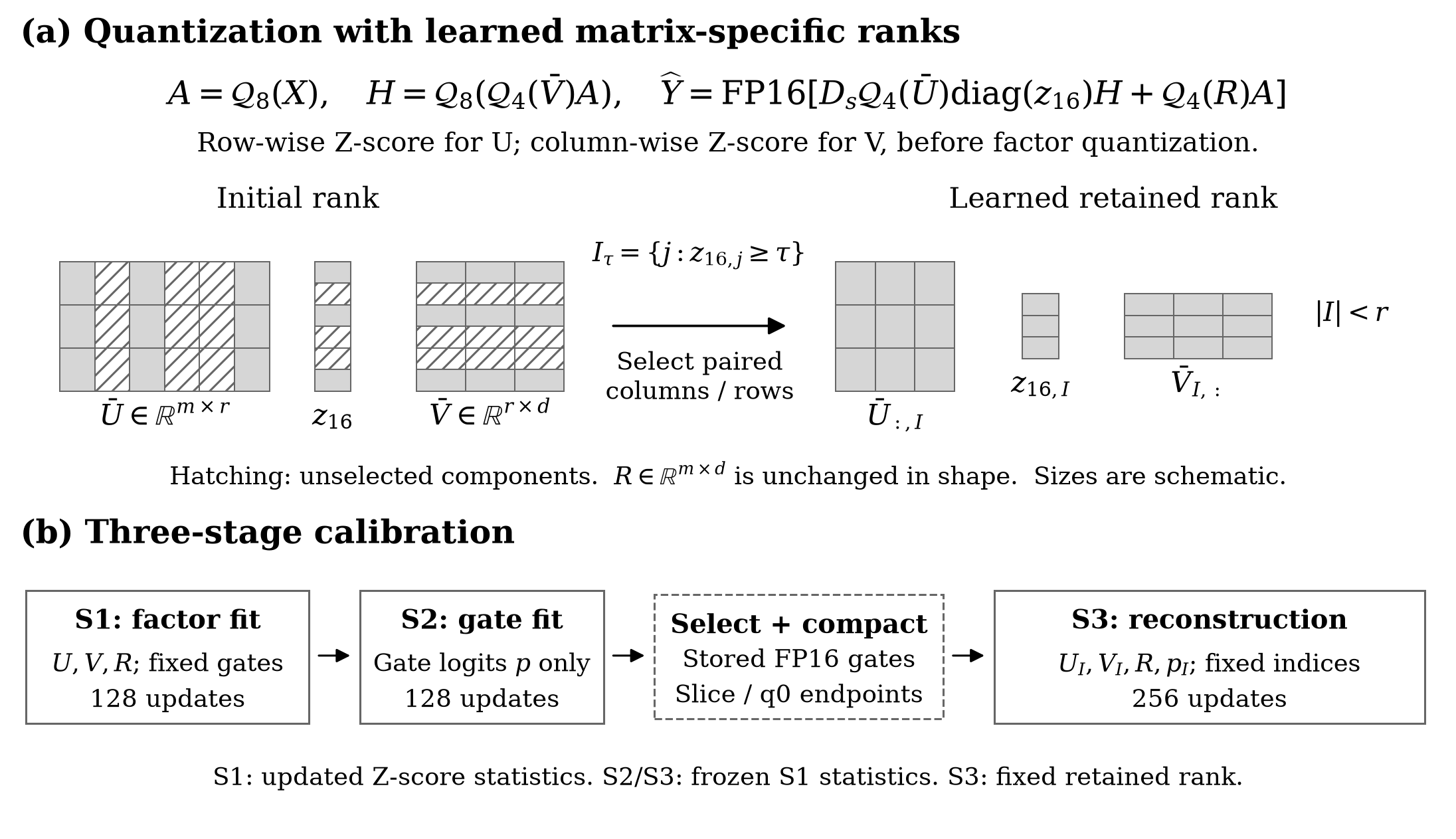}
\caption{GateTrim quantization with learned matrix-specific ranks. Factors are standardized by row (U) and column (V) before quantization; continuous gates determine each matrix's retained rank under a shared threshold. Hatching marks unselected components in a schematic example. The dense residual retains its dimensions. S1 updates normalization statistics, while S2 and S3 reuse frozen S1 statistics; S3 fixes the selected rank.}\end{figure}

\FloatBarrier\subsection{Quantized Low-Rank--Residual Representation}

Let $W\in\mathbb R^{m\times d}$ be the original linear-layer weight and $X\in\mathbb R^{d\times n}$ the input, where $m$, $d$, and $n$ denote the numbers of output channels, input channels, and input vectors, respectively. GateTrim approximates the original output $WX$ using a low-rank branch and a dense residual branch. The low-rank branch comprises factors $U\in\mathbb R^{m\times r}$ and $V\in\mathbb R^{r\times d}$, where $r$ is the number of latent directions; the residual $R\in\mathbb R^{m\times d}$ retains the shape of the original weight.

Before factor quantization, each output row of $U$ and each input column of $V$ is Z-score normalized along the latent dimension, yielding $\bar U$ and $\bar V$. Normalization subtracts the mean and divides by the population standard deviation, with the latter floored at $10^{-6}$. The normalization statistics are computed dynamically in Stage 1 and then fixed to those at the selected Stage 1 checkpoint.

To scale the output channels of the low-rank branch, row-scaling coefficients are computed from the mean squared values of the original weight rows and stored in FP16. Denoting the actual stored coefficients by $s_i$, we have

\begin{equation*}s_i=\operatorname{FP16}\!\left[\max\left(\frac{1}{d}\sum_{j=1}^{d}W_{ij}^{2},10^{-8}\right)\right],\qquad D_s=\operatorname{diag}(s_1,\ldots,s_m).\end{equation*}

Here, $s_i$ corresponds to output channel $i$, and $D_s\in\mathbb R^{m\times m}$ is the diagonal matrix formed from these coefficients. This definition uses the mean squared weight value without taking a square root; the converted coefficients must be finite and strictly positive. $D_s$ scales only the rows of the low-rank branch output, which is then added to the residual branch output.

Within the low-rank branch, gates control the amplitudes of individual latent directions. Let $p\in\mathbb R^r$ denote the trainable gate parameters. Applying the sigmoid function and converting to FP16 gives the gate values used in the forward computation:

\begin{equation*}z_{16}=\operatorname{FP16}\bigl(\sigma(p)\bigr).\end{equation*}

The $k$th entry of $z_{16}$ corresponds to latent direction $k$. Thus, the row-scaling coefficient $s_i$ controls the amplitude of an output channel, whereas the gate value $z_{16,k}$ controls the amplitude of a latent direction; they act along different dimensions.

We use a W4A8 configuration, with $\mathcal Q_4$ and $\mathcal Q_8$ denoting quantization-dequantization operations for four-bit weights and eight-bit activations, respectively. Weight quantization is applied to $\bar U,\bar V,R$ in groups of at most 128 elements within each output row, with group quantization scales stored in FP16. Activation quantization groups features within each input vector, also using groups of at most 128 elements, with dynamically computed scales. The group scales inside the quantizers and the row-scaling coefficients $s$ are defined and applied separately.

The forward computation first quantizes the input, computes the low-rank intermediate activation through the input factor, and quantizes this intermediate activation:

\begin{equation*}A=\mathcal Q_8(X),\qquad H=\mathcal Q_8\!\left(\mathcal Q_4(\bar V)A\right).\end{equation*}

The outputs of the low-rank and residual branches are then added:

\begin{equation*}\widehat Y=\operatorname{FP16}\!\left[D_s\mathcal Q_4(\bar U)\operatorname{diag}(z_{16})H+\mathcal Q_4(R)A\right].\end{equation*}

The first term is the gated and row-scaled low-rank branch output, and the second is the residual branch output. Gates act after weight quantization and outside the intermediate activation quantization operation; they are not folded into $U$ before weight quantization. The row-scaling matrix $D_s$ multiplies only the first term, whereas the outer $\operatorname{FP16}(\cdot)$ casts the sum of the two branches to FP16.

In the current QDQ implementation, the FP16-stored row-scaling coefficients and gates are promoted to FP32 for computation. Branch arithmetic and summation use FP32, followed by conversion to FP16 at the linear-layer output boundary. The residual branch remains dense throughout, so retained width denotes the number of latent directions retained in the low-rank branch. The next subsection defines direction selection using the stored gates and constructs physically compact factors accordingly.

\FloatBarrier\subsection{Threshold-Based Rank Selection}

Every target matrix starts with $r_0=128$. A threshold $\tau$ selects

\begin{equation*}I_\tau=\{j:\operatorname{FP32}(z_{16,j})\geq\tau\}.\end{equation*}

If the set is empty, the largest stored gate is retained, breaking ties by the smallest original index. Indices preserve original order, and $1\leq|I_\tau|\leq128$. Selection uses the actual stored FP16 gates.

For target matrix $\ell$, the learned gates and shared threshold determine its retained rank:

\begin{equation*}r_\ell(\tau)=\max\left(1,\sum_{j=1}^{r_0}\mathbf{1}\{\operatorname{FP32}(z_{16,\ell j})\geq\tau\}\right).\end{equation*}

The gate parameters are optimized by gradients; the integer rank is then derived by this deterministic rule. A common threshold can therefore produce different ranks for different matrices.

For a common parent, the sets are nested as the threshold increases. We study $\tau\in\{0.3,0.5,0.7\}$. Compaction extracts columns of $U$, rows of $V$, and the corresponding gate parameters. Residual and row scales remain present. Stage 3 holds the indices fixed while allowing surviving amplitudes to change; no repeated thresholding or reactivation is performed.

\FloatBarrier\subsection{Three-Stage Calibration and Reconstruction}

Teacher input--output pairs are collected from the original FP16 model for each matrix. Local reconstruction quality is measured by

\begin{equation*}\operatorname{NMSE}=\frac{\sum_k\|\widehat Y_k-Y_k\|_F^2}{\max(\sum_k\|Y_k\|_F^2,10^{-12})}.\end{equation*}

\textbf{Stage 1: representation reconstruction.} Optimize $U,V,R$ for 128 updates with fixed logits and dynamic normalization. Evaluate initialization and every 32 updates on development segments; select the lowest development NMSE, resolving ties in favor of the earlier checkpoint.

\textbf{Stage 2: gate calibration.} Freeze the selected factors, residual, and statistics. Optimize only $p$ for 128 updates. Minibatch MSE is divided by fixed fit-set teacher energy per element and the Stage 1 fit NMSE, with numerical floors. The capacity-penalty coefficient is zero. Use the fixed final update, without development-based checkpoint selection.

\textbf{Stage 3: compact reconstruction.} Independently compact each threshold branch from the Stage 2 parent. Optimize surviving factors, residual, and logits for 256 updates while keeping indices and statistics fixed. Use the same development-checkpoint selection rule as Stage 1.

\par\medskip\noindent\begin{minipage}{\linewidth}\centering\small\setlength{\tabcolsep}{4pt}\renewcommand{\arraystretch}{1.22}

\begin{tabular}{>{\raggedright\arraybackslash}p{\dimexpr 0.10000\linewidth-2\tabcolsep\relax}>{\raggedright\arraybackslash}p{\dimexpr 0.22500\linewidth-2\tabcolsep\relax}>{\raggedright\arraybackslash}p{\dimexpr 0.22500\linewidth-2\tabcolsep\relax}>{\raggedright\arraybackslash}p{\dimexpr 0.22500\linewidth-2\tabcolsep\relax}>{\raggedright\arraybackslash}p{\dimexpr 0.22500\linewidth-2\tabcolsep\relax}}\toprule

\textbf{Stage} & \textbf{Trainable parameters} & \textbf{Updates} & \textbf{Learning rate} & \textbf{Selection} \\ \midrule

S1 & $U,V,R$ & 128 & $10^{-5}$ & Best development NMSE \\

S2 & $p$ & 128 & $10^{-3}$ & Fixed final update \\

S3 & $U_I,V_I,R,p_I$ & 256 per branch & $10^{-5}$ & Best development NMSE \\

\bottomrule\end{tabular}\end{minipage}\par\medskip

Stages 1 and 3 optimize MSE. AdamW uses $\beta=(0.9,0.999)$, $\epsilon=10^{-8}$, zero weight decay, and no gradient clipping. Optimizer state is reset between stages. A minibatch contains four 512-token sequences. Each branch follows a 512-update path per matrix; shared parent stages can be reused across thresholds.

\textbf{Algorithm 1.} Initialize a rank-128 factorization and residual; reconstruct S1 and freeze selected normalization; calibrate gates through S2; for each threshold, select indices from stored gates, construct the slice and q0 endpoints, reconstruct with fixed indices, and export the development-best representation; independently reload and verify the export.

\FloatBarrier\subsection{Compact Export and Storage Accounting}

We distinguish \textbf{slice}, which preserves inherited weight codes, scales, and group mapping after rank reduction; \textbf{q0}, which requantizes the compact factors from the same parent without optimization; and \textbf{post}, which reconstructs q0 with Stage 3. Slice-to-q0 measures compact requantization, whereas q0-to-post measures reconstruction at fixed dimensions.

Four-bit codes are packed at the actual compact shape, without padding the rank back to 128. Scales, gates, row factors, indices, grouping metadata, and container overhead are included in measured files. We report

\begin{equation*}B_{\mathrm{target}}=\sum_\ell\operatorname{filesize}(\mathrm{export}_\ell),\qquad B_{\mathrm{component}}=B_{\mathrm{target}}+B_{\mathrm{non\text{-}target,FP16}}.\end{equation*}

The first quantity is measured; the second estimates complete weight components, counting shared parameters once. It is not the measured size of a separately serialized complete checkpoint. Reload checks establish output consistency within the implementation, rather than independent scientific replication.

\FloatBarrier\section{Experiments}

\FloatBarrier\subsection{Experimental Setup}

\textbf{Models.} Experiments cover Qwen3-4B/8B/14B, Llama-3.2-3B-Instruct, and Llama-3.1-8B-Instruct. All five models have a complete three-threshold evaluation. Each layer contributes seven targets: attention query/key/value/output and MLP gate/up/down projections. Model-family context is provided by the Qwen3 report~\cite{ref9}, Llama 3 report~\cite{ref10}, and the Llama-3.2-3B-Instruct model card~\cite{ref11}.

\par\medskip\noindent\begin{minipage}{\linewidth}\centering\small\setlength{\tabcolsep}{4pt}\renewcommand{\arraystretch}{1.22}

\begin{tabular}{>{\raggedright\arraybackslash}p{\dimexpr 0.34000\linewidth-2\tabcolsep\relax}>{\raggedright\arraybackslash}p{\dimexpr 0.22000\linewidth-2\tabcolsep\relax}>{\raggedright\arraybackslash}p{\dimexpr 0.22000\linewidth-2\tabcolsep\relax}>{\raggedright\arraybackslash}p{\dimexpr 0.22000\linewidth-2\tabcolsep\relax}}\toprule

\textbf{Model} & \textbf{Layers} & \textbf{Target matrices} & \textbf{Quantized endpoints} \\ \midrule

Qwen3-4B & 36 & 252 & Complete \\

Qwen3-8B & 36 & 252 & Complete \\

Qwen3-14B & 40 & 280 & Complete \\

Llama-3.2-3B-Instruct & 28 & 196 & Complete \\

Llama-3.1-8B-Instruct & 32 & 224 & Complete \\

\bottomrule\end{tabular}\end{minipage}\par\medskip

\textbf{Calibration.} Eight fit segments and eight separate development segments from WikiText-2 train contain 512 tokens each: 4,096 fit and 4,096 development tokens. Development samples receive no gradients. Targets use original-model teacher activations, rather than end-to-end language-model training.

\textbf{Evaluation.} WikiText-2 raw test rows are joined with two newline characters and tokenized without a chat template or added special tokens. Non-overlapping 2,048-token windows use batch size one and discard the incomplete tail. Each window contributes 2,047 labels. Cross-entropy is FP32 and NLL accumulation FP64:

\begin{equation*}\operatorname{PPL}=\exp\left(\frac{\sum_{t\in\mathcal T_{\mathrm{valid}}}-\log p(x_t\mid x_{<t})}{|\mathcal T_{\mathrm{valid}}|}\right).\end{equation*}

Qwen evaluations have 146 windows/298,862 labels; Llama evaluations have 141 windows/288,627 labels. Comparisons are within a checkpoint and tokenizer. WikiText source~\cite{ref8}.

\textbf{Initialization seeds.} The main comparison across all five models uses initialization seed 0. For Qwen3-8B, we additionally evaluate initialization seed 1 for all three thresholds. Calibration sampling and the evaluation protocol are held fixed across the two initialization seeds.

\textbf{Implementation.} The reference implementation uses PyTorch~\cite{ref12} and Transformers~\cite{ref13} and executes floating-point QDQ. Audited preflight/identity records list Python 3.12.3, PyTorch 2.8.0, Transformers 5.17.0, and Tokenizers 0.23.2; Appendix A specifies their provenance.

This is an exploratory study. Individual candidate sets were frozen before their corresponding evaluations, but the research process had previous exposure to this test benchmark.

\FloatBarrier\subsection{Main Perplexity Results}

\par\medskip\noindent\begin{minipage}{\linewidth}\centering\small\setlength{\tabcolsep}{4pt}\renewcommand{\arraystretch}{1.22}

\captionof*{table}{\textbf{Table 1. Token perplexity, initialization seed 0.} Bold indicates the best of the three quantized thresholds; lower is better.}

\begin{tabular}{>{\raggedright\arraybackslash}p{\dimexpr 0.34000\linewidth-2\tabcolsep\relax}>{\raggedright\arraybackslash}p{\dimexpr 0.16500\linewidth-2\tabcolsep\relax}>{\raggedright\arraybackslash}p{\dimexpr 0.16500\linewidth-2\tabcolsep\relax}>{\raggedright\arraybackslash}p{\dimexpr 0.16500\linewidth-2\tabcolsep\relax}>{\raggedright\arraybackslash}p{\dimexpr 0.16500\linewidth-2\tabcolsep\relax}}\toprule

\textbf{Model} & \textbf{FP16} & \textbf{$\tau=0.3$} & \textbf{$\tau=0.5$} & \textbf{$\tau=0.7$} \\ \midrule

Qwen3-4B & 13.638088 & 16.247287 & 16.194642 & \textbf{15.407381} \\

Qwen3-8B & 9.715166 & 11.706547 & 11.683825 & \textbf{11.244829} \\

Qwen3-14B & 8.638093 & 9.726787 & 9.663617 & \textbf{9.490028} \\

Llama-3.2-3B-Instruct & 11.046979 & 12.870684 & 12.487662 & \textbf{12.485824} \\

Llama-3.1-8B-Instruct & 7.211869 & 7.908503 & 7.922243 & \textbf{7.861608} \\

\bottomrule\end{tabular}\end{minipage}\par\medskip

Threshold 0.7 gives the lowest PPL among the tested thresholds for all five models. Relative to threshold 0.3, the reduction ranges from 0.59\% to 5.17\%. The response is not uniformly monotonic: threshold 0.5 slightly worsens Llama-3.1-8B-Instruct relative to 0.3. The Llama-3.2-3B-Instruct difference between 0.5 and 0.7 is only 0.001839 PPL.

All quantized settings remain worse than FP16. Even the best tested threshold raises PPL by approximately 9.01\%--15.75\%. These results establish a trade-off within the quantized representation, not FP16-level quality preservation.

\begin{figure}[!htbp]\centering
\includegraphics[width=0.82\linewidth,height=0.82\textheight,keepaspectratio]{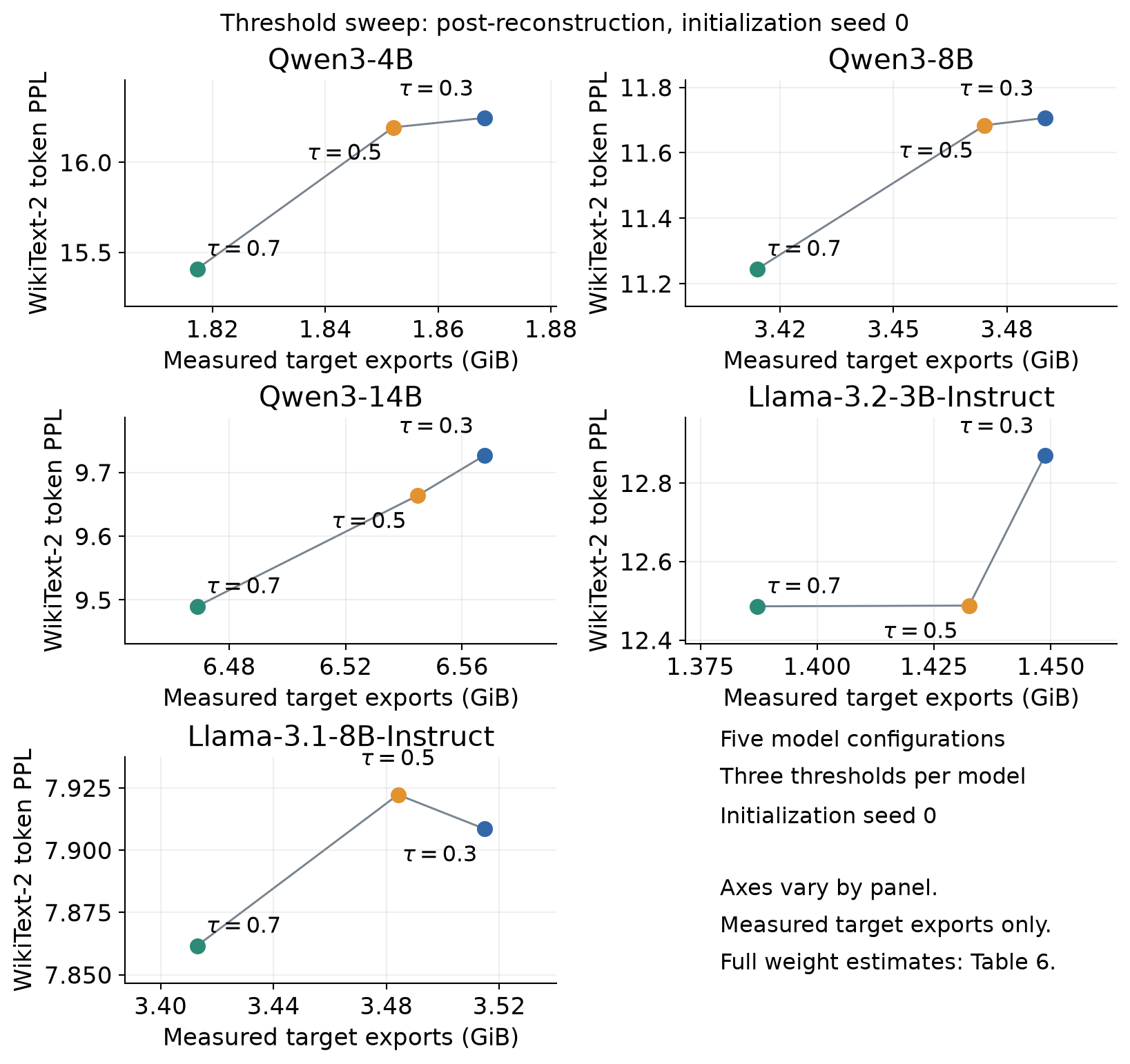}
\caption{The complete three-threshold sweep. Axes are panel-specific; points compare configurations within a model. Lines connect measured settings and do not represent interpolation experiments.}\end{figure}

\FloatBarrier\subsection{Comparison with Published TwinQuant Results}

Table 2 compares GateTrim with the published Qwen3 results from TwinQuant. FP16 denotes our measured full-precision baseline. The TwinQuant W4A8 values are quoted directly from Table 5 of its original paper and are not a local reproduction. TwinQuant~\cite{ref1}.

\par\medskip\noindent\begin{minipage}{\linewidth}\centering\small\setlength{\tabcolsep}{4pt}\renewcommand{\arraystretch}{1.22}

\captionof*{table}{\textbf{Table 2. Qwen3 WikiText-2 PPL with published TwinQuant results [1, Table 5].} FP16 is our measured baseline. GateTrim reports seed-0 post-reconstruction W4A8 at $\tau=0.7$. Bold identifies the lower reported W4A8 value; this is a cross-study numerical reference.}

\begin{tabular}{>{\raggedright\arraybackslash}p{\dimexpr 0.34000\linewidth-2\tabcolsep\relax}>{\raggedright\arraybackslash}p{\dimexpr 0.22000\linewidth-2\tabcolsep\relax}>{\raggedright\arraybackslash}p{\dimexpr 0.22000\linewidth-2\tabcolsep\relax}>{\raggedright\arraybackslash}p{\dimexpr 0.22000\linewidth-2\tabcolsep\relax}}\toprule

\textbf{Model} & \textbf{FP16} & \textbf{TwinQuant W4A8 [1]} & \textbf{GateTrim W4A8} \\ \midrule

Qwen3-4B & 13.638088 & 18.3 & \textbf{15.407381} \\

Qwen3-8B & 9.715166 & 11.5 & \textbf{11.244829} \\

Qwen3-14B & 8.638093 & 10.5 & \textbf{9.490028} \\

\bottomrule\end{tabular}\end{minipage}\par\medskip

GateTrim's measured PPL values are numerically lower in these three configurations. However, checkpoint identity, calibration conditions, and the full evaluation protocols have not been matched. These cross-study observations therefore do not establish controlled superiority. Threshold 0.7 is the best of the three thresholds evaluated in this exploratory study, rather than a prespecified universal optimum; Table 1 retains the complete sweep.

TwinQuant uses 128 calibration prompts of 2,048 tokens, whereas our fit and separate development pools each contain eight 512-token segments. The source does not fully bind the PPL denominator, evaluation-window rules, and checkpoint revisions. Calibration sequence length must not be treated as verified evaluation context. Appendix F records these differences. This table supports no head-to-head claim about speed, memory, storage, or calibration efficiency.

The TwinQuant reference is restricted to Qwen3. Local Llama threshold results are reported in the main experiments.

\FloatBarrier\subsection{Threshold Sensitivity and Retained Rank}

\par\medskip\noindent\begin{minipage}{\linewidth}\centering\small\setlength{\tabcolsep}{4pt}\renewcommand{\arraystretch}{1.22}

\captionof*{table}{\textbf{Table 3. Unweighted mean retained ranks across target matrices.}}

\begin{tabular}{>{\raggedright\arraybackslash}p{\dimexpr 0.34000\linewidth-2\tabcolsep\relax}>{\raggedright\arraybackslash}p{\dimexpr 0.16500\linewidth-2\tabcolsep\relax}>{\raggedright\arraybackslash}p{\dimexpr 0.16500\linewidth-2\tabcolsep\relax}>{\raggedright\arraybackslash}p{\dimexpr 0.16500\linewidth-2\tabcolsep\relax}>{\raggedright\arraybackslash}p{\dimexpr 0.16500\linewidth-2\tabcolsep\relax}}\toprule

\textbf{Model} & \textbf{Mean at 0.3} & \textbf{Mean at 0.5} & \textbf{Mean at 0.7} & \textbf{Median at 0.7} \\ \midrule

Qwen3-4B & 121.22 & 103.24 & 64.73 & 56.5 \\

Qwen3-8B & 116.98 & 102.63 & 55.21 & 43.5 \\

Qwen3-14B & 116.47 & 101.68 & 59.07 & 47.0 \\

Llama-3.2-3B-Instruct & 125.65 & 99.92 & 40.34 & 26.0 \\

Llama-3.1-8B-Instruct & 125.89 & 98.12 & 40.85 & 23.0 \\

\bottomrule\end{tabular}\end{minipage}\par\medskip

Threshold 0.3 preserves most initial components. At 0.7, retained ranks range from 1 to 128 in every completed model. Mean--median differences, particularly for Llama, show that some matrices retain considerably more capacity than a typical matrix. A common threshold therefore produces heterogeneous allocation. Appendix G (Figure 7) shows the distributions across matrices at all three thresholds.

\begin{figure}[!htbp]\centering
\includegraphics[width=\linewidth,height=0.82\textheight,keepaspectratio]{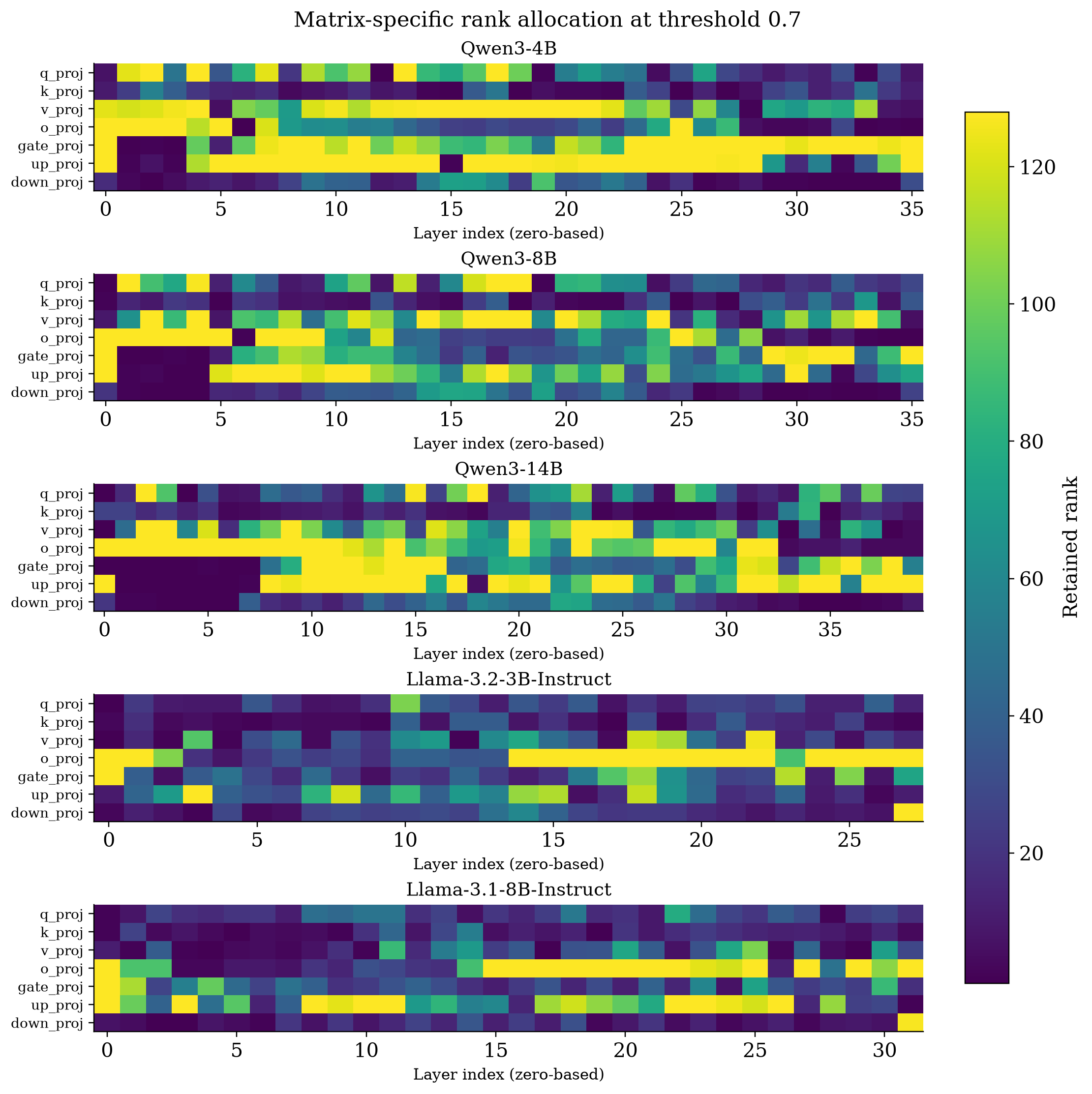}
\caption{Retained rank at threshold 0.7 for every target projection. Layer indices are zero-based. All heatmaps use the same 1--128 scale; no missing values are imputed.}\end{figure}

Reduced rank and improved PPL are consistent with useful capacity allocation after reconstruction, but they do not isolate that mechanism from optimization or regrouping effects. The mean retained rank is not storage-weighted and must not be read as the compression ratio of the complete model.

\FloatBarrier\subsection{Compaction and Reconstruction Ablation}

\par\medskip\noindent\begin{minipage}{\linewidth}\centering\small\setlength{\tabcolsep}{4pt}\renewcommand{\arraystretch}{1.22}

\captionof*{table}{\textbf{Table 4. Qwen3-8B seed-0 endpoints.} Storage is the measured sum of 252 target files; slice/q0/post share dimensions within a threshold.}

\begin{tabular}{>{\raggedright\arraybackslash}p{\dimexpr 0.34000\linewidth-2\tabcolsep\relax}>{\raggedright\arraybackslash}p{\dimexpr 0.16500\linewidth-2\tabcolsep\relax}>{\raggedright\arraybackslash}p{\dimexpr 0.16500\linewidth-2\tabcolsep\relax}>{\raggedright\arraybackslash}p{\dimexpr 0.16500\linewidth-2\tabcolsep\relax}>{\raggedright\arraybackslash}p{\dimexpr 0.16500\linewidth-2\tabcolsep\relax}}\toprule

\textbf{Configuration} & \textbf{Slice PPL} & \textbf{q0 PPL} & \textbf{Post PPL} & \textbf{Target GiB} \\ \midrule

Threshold 0.3 & 12.558078 & 12.552385 & 11.706547 & 3.490113 \\

Threshold 0.5 & 12.523200 & 12.502614 & 11.683825 & 3.474115 \\

Threshold 0.7 & 11.646183 & 11.624194 & 11.244829 & 3.413956 \\

\bottomrule\end{tabular}\end{minipage}\par\medskip

Compact requantization gives small PPL reductions relative to inherited-code slicing; reconstruction gives larger reductions. For Qwen3-8B at threshold 0.7, reconstruction after rank selection reduces perplexity from 11.6242 at the compacted and requantized q0 endpoint to 11.2448 at the post endpoint, a relative reduction of approximately 3.26\%, with unchanged retained dimensions and measured target-export size. This improvement is measured relative to q0. The final perplexity remains approximately 15.75\% above the FP16 baseline of 9.7152.

\begin{figure}[!htbp]\centering
\includegraphics[width=\linewidth,height=0.82\textheight,keepaspectratio]{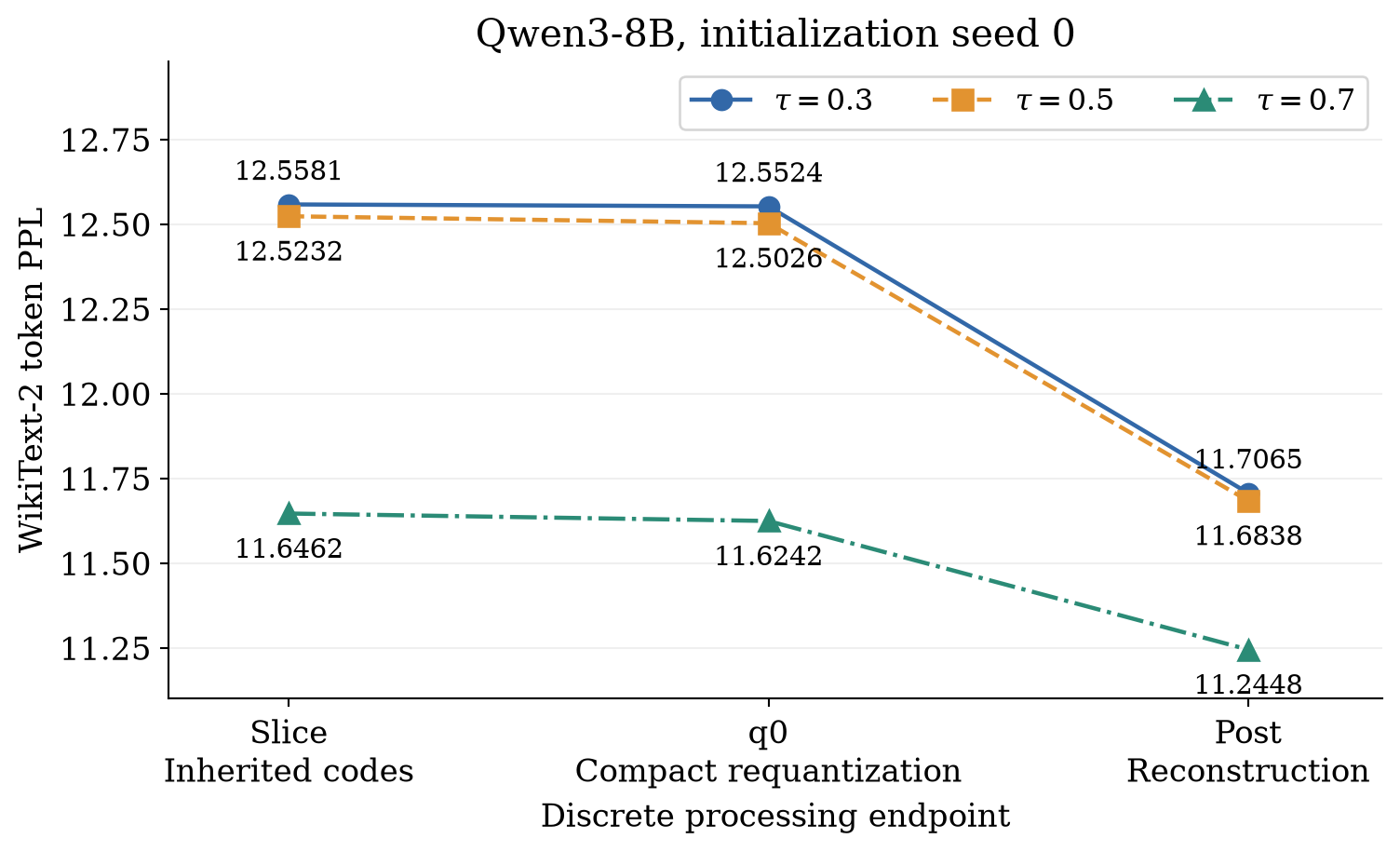}
\caption{Qwen3-8B seed-0 PPL at the slice, q0, and post endpoints for each threshold. Lines connect three discrete processing states; they are not optimizer convergence traces. Within each threshold, target dimensions and measured export bytes are unchanged. Table 4 gives the complete values.}\end{figure}

\FloatBarrier\subsection{Local Reconstruction Error}

We compare each reconstructed matrix with its own q0 endpoint at the same retained rank. A matrix meets the reported improvement criterion when its development NMSE decreases by at least 0.1\%. This local criterion is reported separately from whole-model perplexity.

\par\medskip\noindent\begin{minipage}{\linewidth}\centering\small\setlength{\tabcolsep}{4pt}\renewcommand{\arraystretch}{1.22}

\captionof*{table}{\textbf{Table 5. Qwen3-8B q0-to-post development-NMSE improvement at threshold 0.7.}}

\begin{tabular}{>{\raggedright\arraybackslash}p{\dimexpr 0.34000\linewidth-2\tabcolsep\relax}>{\raggedright\arraybackslash}p{\dimexpr 0.22000\linewidth-2\tabcolsep\relax}>{\raggedright\arraybackslash}p{\dimexpr 0.22000\linewidth-2\tabcolsep\relax}>{\raggedright\arraybackslash}p{\dimexpr 0.22000\linewidth-2\tabcolsep\relax}}\toprule

\textbf{Initialization seed} & \textbf{Target matrices} & \textbf{Improved by at least 0.1\%} & \textbf{Below the improvement criterion} \\ \midrule

0 & 252 & 249 & 3 \\

1 & 252 & 249 & 3 \\

\bottomrule\end{tabular}\end{minipage}\par\medskip

\begin{figure}[!htbp]\centering
\includegraphics[width=\linewidth,height=0.82\textheight,keepaspectratio]{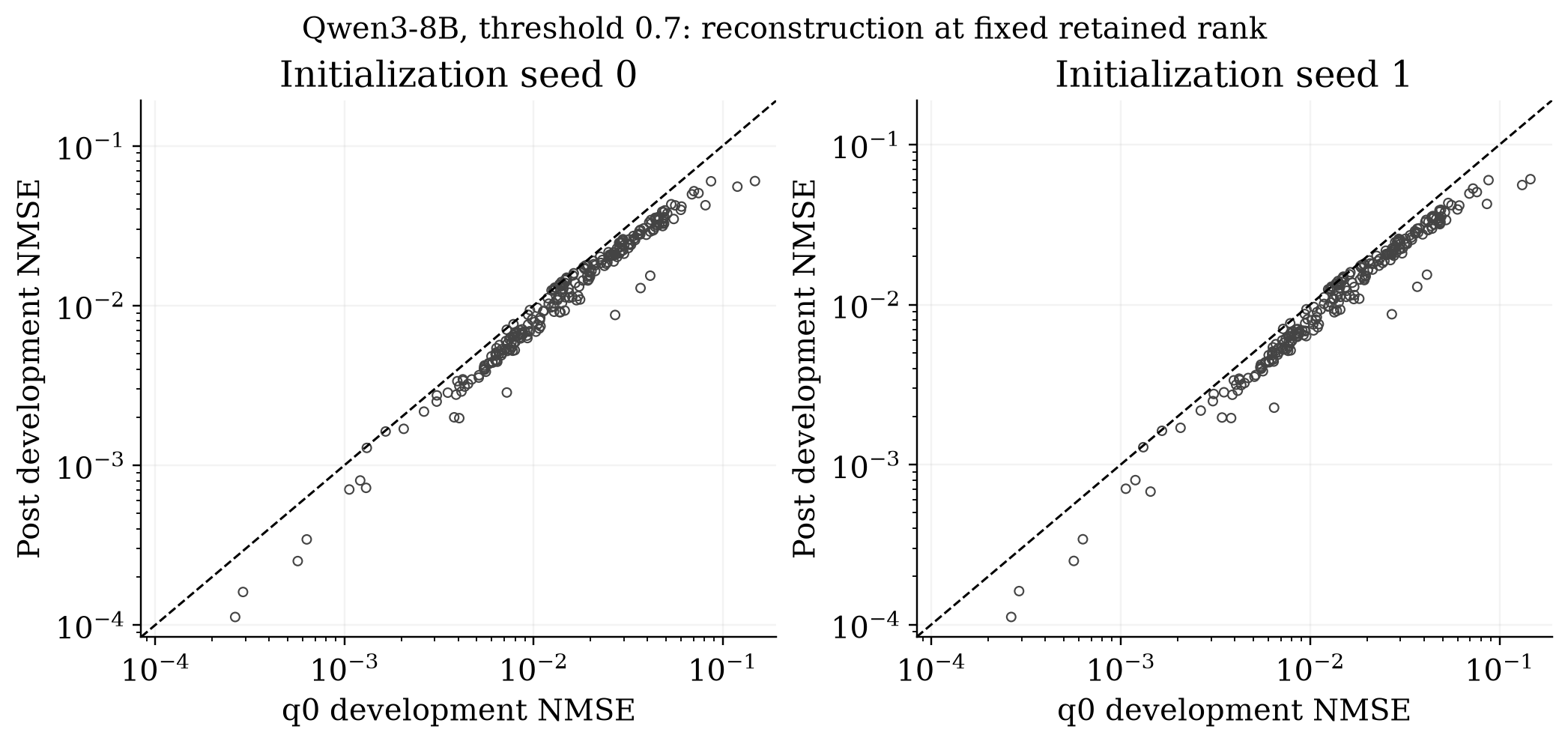}
\caption{Development NMSE before (q0) and after (post) reconstruction for all 252 Qwen3-8B target matrices at threshold 0.7, shown separately for initialization seeds 0 and 1. The dashed diagonal marks equality; points below it have lower NMSE after reconstruction. Both axes use logarithmic scales.}\end{figure}

For Qwen3-4B and Llama-3.2-3B-Instruct, all targets satisfy the 0.1\% improvement criterion at every threshold. Llama-3.1-8B-Instruct satisfies it for 224/224 matrices at 0.3 and 0.5, and 223/224 at 0.7. The remaining matrix, \texttt{model.\allowbreak{}layers.\allowbreak{}1.\allowbreak{}mlp.\allowbreak{}down\_\allowbreak{}proj}, has unchanged development NMSE of 0.000980936632. For Qwen3-14B, the corresponding improvement counts are 277/280, 273/280, and 270/280 for thresholds 0.3, 0.5, and 0.7. Improved whole-model PPL therefore does not imply that every matrix meets this local improvement criterion.

\FloatBarrier\subsection{Storage, Runtime Memory, and Calibration Cost}

\par\medskip\noindent\begin{minipage}{\linewidth}\centering\small\setlength{\tabcolsep}{4pt}\renewcommand{\arraystretch}{1.22}

\captionof*{table}{\textbf{Table 6. Measured target exports and estimated complete weight components, GiB.}}

\begin{tabular}{>{\raggedright\arraybackslash}p{\dimexpr 0.34000\linewidth-2\tabcolsep\relax}>{\raggedright\arraybackslash}p{\dimexpr 0.16500\linewidth-2\tabcolsep\relax}>{\raggedright\arraybackslash}p{\dimexpr 0.16500\linewidth-2\tabcolsep\relax}>{\raggedright\arraybackslash}p{\dimexpr 0.16500\linewidth-2\tabcolsep\relax}>{\raggedright\arraybackslash}p{\dimexpr 0.16500\linewidth-2\tabcolsep\relax}}\toprule

\textbf{Model} & \textbf{Target, 0.3} & \textbf{Target, 0.5} & \textbf{Target, 0.7} & \textbf{Complete components, 0.7, estimate} \\ \midrule

Qwen3-4B & 1.868268 & 1.852043 & 1.817271 & 2.542124 \\

Qwen3-8B & 3.490113 & 3.474115 & 3.413956 & 5.732889 \\

Qwen3-14B & 6.567802 & 6.544708 & 6.468959 & 9.367700 \\

Llama-3.2-3B-Instruct & 1.448747 & 1.432495 & 1.387206 & 2.121419 \\

Llama-3.1-8B-Instruct & 3.514891 & 3.484129 & 3.412931 & 5.370458 \\

\bottomrule\end{tabular}\end{minipage}\par\medskip

Moving from threshold 0.3 to 0.7 reduces measured target-file storage by approximately 1.50\%--4.25\%. Component accounting explains the limited reduction: at 0.7, residual codes and scales occupy 96.01\%, 97.70\%, 98.08\%, 97.57\%, and 98.20\% of target-file bytes for Qwen3-4B, Qwen3-8B, Qwen3-14B, Llama-3.2-3B-Instruct, and Llama-3.1-8B-Instruct, respectively.

\begin{figure}[!htbp]\centering
\includegraphics[width=\linewidth,height=0.82\textheight,keepaspectratio]{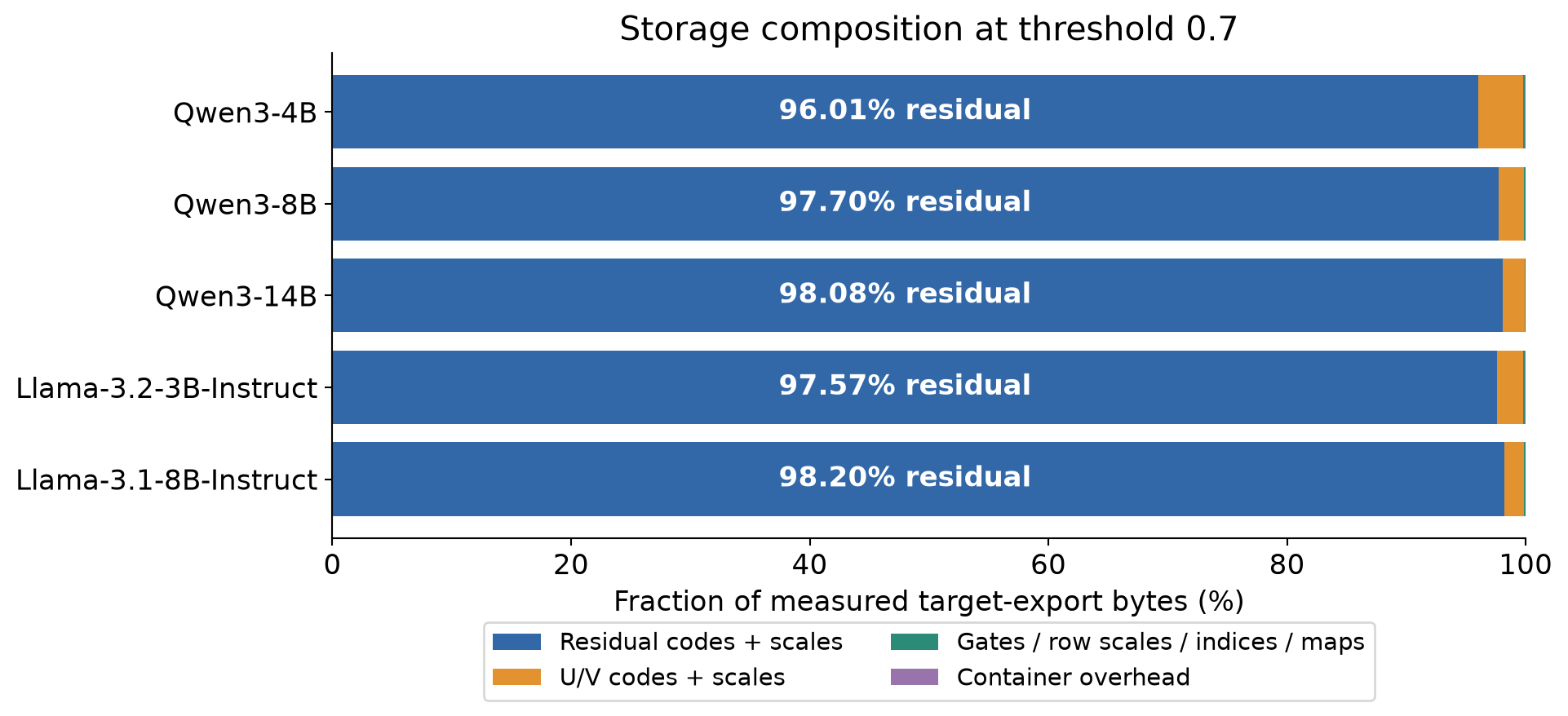}
\caption{Exact decomposition of measured target-file bytes at threshold 0.7. The denominator excludes non-target model parameters; it is not complete-checkpoint storage. Detailed code, scale, and metadata accounting is retained in the authors' records.}\end{figure}

Packed parameters are decoded into floating-point buffers in the reference implementation. Export savings therefore do not imply runtime-memory savings.

\par\medskip\noindent\begin{minipage}{\linewidth}\centering\small\setlength{\tabcolsep}{4pt}\renewcommand{\arraystretch}{1.22}

\captionof*{table}{\textbf{Table 7. Resident memory, GiB.} Fixed-forward values are absolute PyTorch allocated-memory peaks including the loaded model, batch one, and 2,048 tokens. Sampled process peaks include loading, assembly, and evaluation.}

\begin{tabular}{>{\raggedright\arraybackslash}p{\dimexpr 0.34000\linewidth-2\tabcolsep\relax}>{\raggedright\arraybackslash}p{\dimexpr 0.16500\linewidth-2\tabcolsep\relax}>{\raggedright\arraybackslash}p{\dimexpr 0.16500\linewidth-2\tabcolsep\relax}>{\raggedright\arraybackslash}p{\dimexpr 0.16500\linewidth-2\tabcolsep\relax}>{\raggedright\arraybackslash}p{\dimexpr 0.16500\linewidth-2\tabcolsep\relax}}\toprule

\textbf{Model} & \textbf{FP16 fixed-forward} & \textbf{0.7 fixed-forward} & \textbf{FP16 process peak} & \textbf{0.7 process peak} \\ \midrule

Qwen3-4B & 8.090 & 16.938 & 12.400 & 20.701 \\

Llama-3.2-3B-Instruct & 6.495 & 13.205 & 10.164 & 16.291 \\

Llama-3.1-8B-Instruct & 15.471 & 30.138 & 19.287 & 34.205 \\

\bottomrule\end{tabular}\end{minipage}\par\medskip

The resident QDQ variants use more memory than FP16. Qwen3-8B threshold 0.7 separately records a sampled evaluation-process peak of 35.430 GiB; that value does not substitute for a matched fixed-forward measurement. A separate Qwen3-4B offload check reduces GPU memory while preserving evaluated outputs, but this is a placement effect (Appendix D). Under the matched Qwen3-14B placement of 32 CPU layers and 8 GPU layers, fixed-forward allocated peaks are 9.857 GiB for FP16 and 14.961 GiB for threshold 0.7. QDQ therefore also uses more GPU memory in this layout; these values are not resident-model measurements.

The recorded main-pipeline occupancies are approximately 2.450 hours for Qwen3-4B and 5.576 hours for both Llama models combined. These are controller-accounted GPU-worker durations, including pipeline operations, not hardware-utilization measurements. Qwen3-14B requires 13.122 cumulative GPU-worker hours, including its FP16 baseline and interrupted attempts. Appendix D separates phases and does not rank unequal run scopes by their totals.

\FloatBarrier\subsection{Additional Initialization Seeds}

Qwen3-8B is evaluated with initialization seeds 0 and 1. In both runs, threshold 0.7 gives the lowest perplexity among the three threshold configurations, reaching 11.244829 and 11.244791, respectively (Appendix C). The preference for threshold 0.7 therefore persists across the two tested initializations, although other configurations show larger differences between seeds. This comparison examines initialization sensitivity under fixed calibration sampling; it does not assess sensitivity to calibration-sample selection or establish statistical significance.

\FloatBarrier\section{Limitations and Future Work}

This initial study evaluates W4A8 only. GateTrim is not defined by that bit allocation, but its quality at other weight or activation precisions remains untested. The current experiments evaluate the combined standardization, gate-learning, rank-selection, and reconstruction procedure; they do not isolate the independent contribution of every component.

Completed model-level results cover WikiText-2 PPL, a benchmark consulted during development. They do not establish transfer to an independent text distribution, preservation of downstream task accuracy, or instruction-following quality. The five-model main comparison uses initialization seed 0, with a second initialization evaluated only on Qwen3-8B. Broader replication across models and independently sampled calibration sets remains necessary to characterize variability.

Each matrix is calibrated against original-model activations, whereas the composed quantized model receives activations from preceding quantized components. This mismatch may contribute to accumulated error. Lower final PPL also does not imply that every matrix meets the local q0-to-post improvement criterion in Section 5.6.

The numerical implementation exports low-bit representations but uses floating-point QDQ. Native compact execution and matched end-to-end measurements are needed before claiming deployment speed or memory advantages. Independent-data evaluation, downstream tasks, and other bit configurations remain future work.

\FloatBarrier\section{Conclusion}

We introduced GateTrim, an adaptive-rank post-training quantization method that learns matrix-specific ranks through gate optimization and a shared threshold. Factor standardization and learned gates support rank determination for each target matrix, followed by reconstruction of its compact quantized representation. In the W4A8 evaluation across five models, threshold 0.7 achieves a lower mean retained rank and lower WikiText-2 perplexity than threshold 0.3, and the lowest values among the three tested thresholds. Qwen3-8B exhibits this threshold preference before reconstruction under both tested initialization seeds; reconstruction provides further improvement without increasing retained rank or export size.

These results demonstrate the reported rank--perplexity relationship within the evaluated GateTrim configurations. Measured export savings, local reconstruction error, and runtime efficiency remain distinct outcomes; the current floating-point QDQ implementation does not establish native low-bit speed or memory benefits.

\FloatBarrier

\appendix

\FloatBarrier\section{Model Identities and Reproducibility}

\par\medskip\noindent\begin{minipage}{\linewidth}\centering\small\setlength{\tabcolsep}{4pt}\renewcommand{\arraystretch}{1.22}

\begin{tabular}{>{\raggedright\arraybackslash}p{\dimexpr 0.46000\linewidth-2\tabcolsep\relax}>{\raggedright\arraybackslash}p{\dimexpr 0.54000\linewidth-2\tabcolsep\relax}}\toprule

\textbf{Repository} & \textbf{Pinned revision} \\ \midrule

\texttt{Qwen/\allowbreak{}Qwen3-4B} & \texttt{1cfa9\allowbreak{}a7208\allowbreak{}91212\allowbreak{}64592\allowbreak{}14e8b\allowbreak{}04321\allowbreak{}603b3\allowbreak{}df60c} \\

\texttt{Qwen/\allowbreak{}Qwen3-8B} & \texttt{b9688\allowbreak{}26d9c\allowbreak{}46dd6\allowbreak{}066d1\allowbreak{}09eab\allowbreak{}c6255\allowbreak{}188de\allowbreak{}91218} \\

\texttt{Qwen/\allowbreak{}Qwen3-14B} & \texttt{40c06\allowbreak{}9824f\allowbreak{}4251a\allowbreak{}91eef\allowbreak{}af281\allowbreak{}ebe4c\allowbreak{}544ef\allowbreak{}d3e18} \\

\texttt{meta-llama/\allowbreak{}Llama-3.\allowbreak{}2-3B-Instruct} & \texttt{0cb88\allowbreak{}a4f76\allowbreak{}4b7a1\allowbreak{}2671c\allowbreak{}53f08\allowbreak{}38cd8\allowbreak{}31a08\allowbreak{}43b95} \\

\texttt{meta-llama/\allowbreak{}Llama-3.\allowbreak{}1-8B-Instruct} & \texttt{0e9e3\allowbreak{}9f249\allowbreak{}a1697\allowbreak{}6918f\allowbreak{}6564b\allowbreak{}8830b\allowbreak{}c894c\allowbreak{}89659} \\

\bottomrule\end{tabular}\end{minipage}\par\medskip

Dataset: \texttt{Salesforce/\allowbreak{}wikitext}, \texttt{wikitext-2-raw-v1}, revision \texttt{b0860\allowbreak{}1e043\allowbreak{}26c79\allowbreak{}dfdd3\allowbreak{}2d625\allowbreak{}aee71\allowbreak{}d232d\allowbreak{}685c3}. The 4,358 test rows are joined by \texttt{\textbackslash{}n\textbackslash{}n}. SDPA is used; KV caching and TF32 are disabled. PPL is computed from total valid-token NLL, not an average of window PPLs or a word-level denominator. FP16 denotes forward computation, not necessarily original checkpoint serialization dtype.

Existing Qwen3-4B, Qwen3-8B reference preflight, and Llama identity records agree on Python 3.12.3, PyTorch 2.8.0, Transformers 5.17.0, Tokenizers 0.23.2, NumPy 2.5.3, Datasets 5.0.1, and PyArrow 25.0.1. Qwen3-4B and Llama runtime manifests further record \texttt{torch 2.\allowbreak{}8.\allowbreak{}0+cu128} and CUDA build 12.8. Llama identity records include Safetensors 0.8.0. The Llama admission snapshot identifies an NVIDIA RTX A6000. A per-run driver-version record and complete hardware/software matrix for all historical runs remain to be consolidated; build CUDA is not the installed driver version. Model-config \texttt{transformers\_\allowbreak{}version} fields are not treated as runtime versions.

\FloatBarrier\section{Initialization and Numerical Details}

The initial residual is a groupwise round-to-nearest QDQ approximation of the original weight. Its remaining error is divided by the original FP32 row factors and decomposed by randomized rank-128 SVD with four iterations. The left factor is scaled by the largest retained singular value (floored at $10^{-8}$); relative singular values, clipped to $[10^{-4},1-10^{-4}]$, initialize the gates, and the right singular vectors initialize $V$. Module-specific seeds are deterministic. Subsequent normalization means the initial normalized operator is not assumed to reproduce the truncated SVD exactly.

Four-bit weights use zero point zero, code range $[-8,7]$, nearest-even rounding, and raw scales $\max(\operatorname{absmax}/7,10^{-12})$ for nonzero groups. Codes use the raw scale, while reconstruction accounts for FP16 scale storage. Zero groups use zero codes and scale one. Finite positive checks and a narrowly scoped floor of $2^{-24}$ for positive scales that otherwise round to FP16 zero are applied consistently.

Activations use range $[-127,127]$, dynamic FP32 scales, and detached scale gradients. Straight-through gradients cover the QDQ operators, stored-gate rounding, and the final FP16 boundary. Normalization statistics remain those selected in Stage 1 after compaction. The rank dimension is never padded back to 128; only a terminal unused nibble may be needed for an odd total element count.

\FloatBarrier\section{Additional Results}

\par\medskip\noindent\begin{minipage}{\linewidth}\centering\small\setlength{\tabcolsep}{4pt}\renewcommand{\arraystretch}{1.22}

\begin{tabular}{>{\raggedright\arraybackslash}p{\dimexpr 0.46000\linewidth-2\tabcolsep\relax}>{\raggedright\arraybackslash}p{\dimexpr 0.27000\linewidth-2\tabcolsep\relax}>{\raggedright\arraybackslash}p{\dimexpr 0.27000\linewidth-2\tabcolsep\relax}}\toprule

\textbf{Configuration} & \textbf{Qwen3-8B seed 0 PPL} & \textbf{Qwen3-8B seed 1 PPL} \\ \midrule

Threshold 0.3 & 11.706547 & 12.089265 \\

Threshold 0.5 & 11.683825 & 11.877027 \\

Threshold 0.7 & 11.244829 & 11.244791 \\

\bottomrule\end{tabular}\end{minipage}\par\medskip

Calibration sampling is fixed. No completed third-seed result is included.

\par\medskip\noindent\begin{minipage}{\linewidth}\centering\small\setlength{\tabcolsep}{4pt}\renewcommand{\arraystretch}{1.22}

\begin{tabular}{>{\raggedright\arraybackslash}p{\dimexpr 0.34000\linewidth-2\tabcolsep\relax}>{\raggedright\arraybackslash}p{\dimexpr 0.22000\linewidth-2\tabcolsep\relax}>{\raggedright\arraybackslash}p{\dimexpr 0.22000\linewidth-2\tabcolsep\relax}>{\raggedright\arraybackslash}p{\dimexpr 0.22000\linewidth-2\tabcolsep\relax}}\toprule

\textbf{Model, threshold 0.7} & \textbf{Rank 1} & \textbf{Rank 128} & \textbf{Total} \\ \midrule

Qwen3-4B & 15 & 56 & 252 \\

Qwen3-8B & 19 & 35 & 252 \\

Qwen3-14B & 29 & 55 & 280 \\

Llama-3.2-3B-Instruct & 3 & 18 & 196 \\

Llama-3.1-8B-Instruct & 7 & 22 & 224 \\

\bottomrule\end{tabular}\end{minipage}\par\medskip

Detailed values and source-file fingerprints are retained in the authors' records. The rank distributions describe variation across target matrices; they do not measure uncertainty across initialization seeds.

\FloatBarrier\section{Resource Measurement and Validation}

Fixed-forward measurements use one warm-up and three measured passes at batch one/2,048 tokens with no KV cache. Allocated, reserved, and process-level peaks are distinct quantities. Process sampling approximately every two seconds can miss short peaks.

Qwen3-4B offload validation preserves all original files and compact exports. All four configurations reproduce the resident evaluation's per-window NLL and PPL; selected layer outputs, logits, and token losses are checked for exact agreement.

\par\medskip\noindent\begin{minipage}{\linewidth}\centering\small\setlength{\tabcolsep}{4pt}\renewcommand{\arraystretch}{1.22}

\begin{tabular}{>{\raggedright\arraybackslash}p{\dimexpr 0.34000\linewidth-2\tabcolsep\relax}>{\raggedright\arraybackslash}p{\dimexpr 0.16500\linewidth-2\tabcolsep\relax}>{\raggedright\arraybackslash}p{\dimexpr 0.16500\linewidth-2\tabcolsep\relax}>{\raggedright\arraybackslash}p{\dimexpr 0.16500\linewidth-2\tabcolsep\relax}>{\raggedright\arraybackslash}p{\dimexpr 0.16500\linewidth-2\tabcolsep\relax}}\toprule

\textbf{Qwen3-4B configuration} & \textbf{Resident allocated GiB} & \textbf{Offloaded allocated GiB} & \textbf{Resident three-forward time, s} & \textbf{Offloaded three-forward time, s} \\ \midrule

FP16 & 8.090242 & 2.761007 & 0.714060 & 12.358798 \\

Threshold 0.7 & 16.937670 & 2.761007 & 4.879314 & 32.955808 \\

\bottomrule\end{tabular}\end{minipage}\par\medskip

The lower GPU footprint is a placement benefit accompanied by more time in this probe, not an algorithmic inference-speed gain.

\par\medskip\noindent\begin{minipage}{\linewidth}\centering\small\setlength{\tabcolsep}{4pt}\renewcommand{\arraystretch}{1.22}

\captionof*{table}{\textbf{Recorded phase durations, hours.} These are elapsed GPU-worker pipeline durations from existing controller stages, including worker operations. They are not isolated kernel time or utilization-adjusted GPU time.}

\begin{tabular}{>{\raggedright\arraybackslash}p{\dimexpr 0.34000\linewidth-2\tabcolsep\relax}>{\raggedright\arraybackslash}p{\dimexpr 0.16500\linewidth-2\tabcolsep\relax}>{\raggedright\arraybackslash}p{\dimexpr 0.16500\linewidth-2\tabcolsep\relax}>{\raggedright\arraybackslash}p{\dimexpr 0.16500\linewidth-2\tabcolsep\relax}>{\raggedright\arraybackslash}p{\dimexpr 0.16500\linewidth-2\tabcolsep\relax}}\toprule

\textbf{Run scope} & \textbf{Teacher h} & \textbf{Train worker h} & \textbf{Reload h} & \textbf{Evaluate h} \\ \midrule

Qwen3-4B, three thresholds & 0.391398 & 1.798801 & 0.035709 & 0.223825 \\

Llama-3.2-3B-Instruct & 0.298013 & 1.374745 & 0.028622 & 0.164483 \\

Llama-3.1-8B-Instruct & 0.541293 & 2.797719 & 0.048457 & 0.321513 \\

\bottomrule\end{tabular}\end{minipage}\par\medskip

The phase table reports runs covering the complete three-threshold pipeline. Tiny admission phases are omitted from the display; detailed historical accounting is retained in the authors' records.

Independent reload validation checks dimensions, packing, scales, selected indices, and canonical outputs. Evaluation summaries are recomputed from saved per-window NLL. These checks establish consistency of artifacts and scoring, not an independently reproduced method. The implementation and supporting records have not yet been publicly released.

\FloatBarrier\subsection{Qwen3-14B Placement and Resource Scope}

All entries below use the fixed 32-CPU-layer/8-GPU-layer layout. Complete weight-component estimates add 3,112,499,200 bytes of non-target FP16 parameters to measured exports. The original weight shards occupy 29,536,665,640 bytes; their original BF16 serialization is unchanged by FP16 evaluation.

\par\medskip\noindent\begin{minipage}{\linewidth}\centering\small\setlength{\tabcolsep}{4pt}\renewcommand{\arraystretch}{1.22}

\begin{tabular}{>{\raggedright\arraybackslash}p{\dimexpr 0.34000\linewidth-2\tabcolsep\relax}>{\raggedright\arraybackslash}p{\dimexpr 0.16500\linewidth-2\tabcolsep\relax}>{\raggedright\arraybackslash}p{\dimexpr 0.16500\linewidth-2\tabcolsep\relax}>{\raggedright\arraybackslash}p{\dimexpr 0.16500\linewidth-2\tabcolsep\relax}>{\raggedright\arraybackslash}p{\dimexpr 0.16500\linewidth-2\tabcolsep\relax}}\toprule

\textbf{Configuration} & \textbf{Fixed allocated GiB} & \textbf{Fixed reserved GiB} & \textbf{Sampled process GiB} & \textbf{Sampled CPU RSS GiB} \\ \midrule

FP16 & 9.856509 & 27.515625 & 27.830078 & 23.713131 \\

Threshold 0.3 & 15.138877 & 27.529297 & 29.599609 & 46.296062 \\

Threshold 0.5 & 15.077720 & 27.529297 & 29.931641 & 46.626961 \\

Threshold 0.7 & 14.960925 & 27.533203 & 29.599609 & 45.827717 \\

\bottomrule\end{tabular}\end{minipage}\par\medskip

FP16 full-test PPL is reused from the accepted 146-window evaluation. Placement validation uses three exact probes, covering 120 layer comparisons, logits, and token losses; it is not another 146-window evaluation. FP16 sampled peaks cover loading, probes, and fixed forwards, whereas quantized sampled peaks also include full PPL evaluation. The fixed-forward columns share the same workload; the sampled columns do not cover identical durations or phases.

The cumulative 47,238.627249 worker seconds include 2,902.056158 seconds for the FP16 baseline and all historical interrupted attempts. The final attempt's recorded phase durations cannot replace that cumulative ledger. All 2,520 exports, including 840 post exports, have accepted file/reload checks.

\FloatBarrier\section{Data and Implementation Record}

The authors retain a local package containing machine-readable measurements, source-file fingerprints, figure-generation scripts, and snapshots of the experiment implementations. These materials support reconstruction of the reported tables and figures without running the models. The snapshots preserve distinct run variants and are not a newly validated, portable training release. Original model weights and training activations are not redistributed. Code and reproducibility materials are being prepared for release and are not yet publicly available. A repository link will be provided when they are released.

The manuscript reports the completed study; independent-data evaluation, downstream tasks, and further seeds are reserved for subsequent work. Their absence limits mechanism and generalization claims but does not change the reported measurements. No additional model experiments were performed in preparing this manuscript version.

\FloatBarrier\section{Scope of the TwinQuant Literature Reference}

\par\medskip\noindent\begin{minipage}{\linewidth}\centering\small\setlength{\tabcolsep}{4pt}\renewcommand{\arraystretch}{1.22}

\captionof*{table}{\textbf{Table 8. Provenance and protocol boundaries for Table 2.} Source: TwinQuant v1, Section 5.1 and Appendix A.3, Table 5 [1]. TwinQuant~\cite{ref1}.}

\begin{tabular}{>{\raggedright\arraybackslash}p{\dimexpr 0.18000\linewidth-2\tabcolsep\relax}>{\raggedright\arraybackslash}p{\dimexpr 0.34000\linewidth-2\tabcolsep\relax}>{\raggedright\arraybackslash}p{\dimexpr 0.48000\linewidth-2\tabcolsep\relax}}\toprule

\textbf{Item} & \textbf{TwinQuant paper} & \textbf{GateTrim in this study} \\ \midrule

Result provenance & Published Qwen3 TwinQuant W4A8 values & Locally measured FP16 and W4A8 QDQ values \\

Model identity & Qwen3 size labels; exact revisions not bound to the table & Official repository IDs and revisions in Appendix A \\

Calibration & 128 prompts, 2,048 tokens each, sampled from WikiText2 & 8 fit + 8 separate development segments, 512 tokens each; no development gradients \\

PPL evaluation & WikiText2 PPL reported; precise denominator and window rules not fully bound & WikiText-2 raw test; token PPL; non-overlapping 2,048-token windows; incomplete tail dropped \\

Input and scoring & Tokenizer/special-token handling and accumulation details not fully bound & No chat template or added special tokens; FP32 cross-entropy, FP64 sum; 146 windows and 298,862 labels per Qwen model \\

Threshold choice & Not applicable to the reported method & Best of the three tested thresholds, 0.7; seed 0; all thresholds in Table 1 \\

Permitted interpretation & Published literature reference & Numerical juxtaposition, not matched algorithm ranking or a system-efficiency comparison \\

\bottomrule\end{tabular}\end{minipage}\par\medskip

A future controlled comparison would fix checkpoint revisions, tokenizer and input tensors, PPL denominator and context rules, calibration samples, and numerical settings before evaluating both methods. It remains deferred work. No TwinQuant reproduction or new evaluation was run for this manuscript revision.

\clearpage\section{Retained-Rank Distributions}

\begin{figure}[!htbp]\centering
\includegraphics[width=\linewidth,height=0.82\textheight,keepaspectratio]{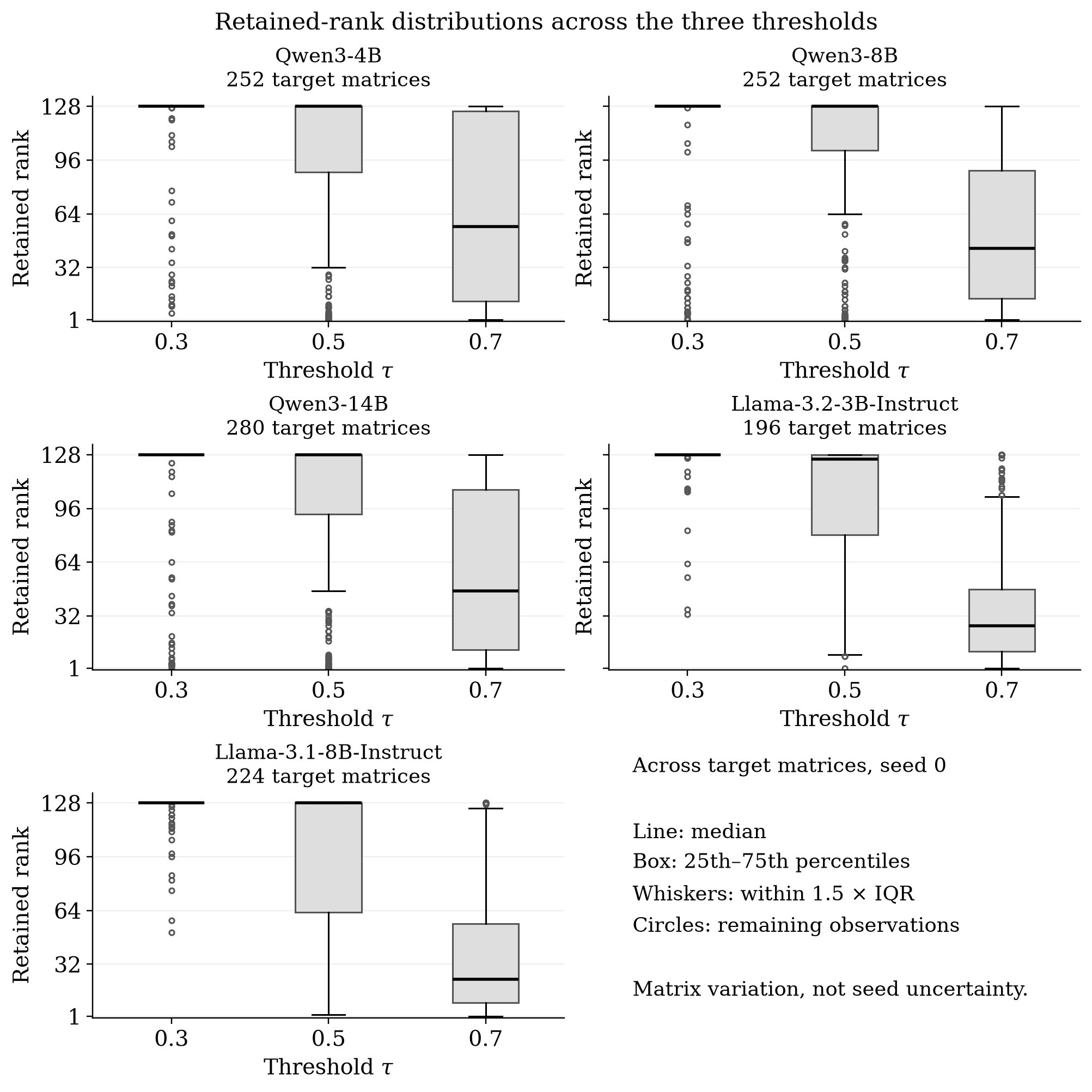}
\caption{Retained ranks across all target matrices at each tested threshold, initialization seed 0. Boxes span the 25th--75th percentiles, center lines show medians, whiskers extend to observed values within 1.5 interquartile ranges, and circles show the remaining observations. Each model uses the same 1--128 rank scale. These distributions describe matrix-to-matrix variation, not uncertainty across seeds. Values are taken from existing post-reconstruction exports; retained indices are fixed during reconstruction.}\end{figure}

The distributions complement the means in Table 3 and the threshold-0.7 heatmaps in Figure 3. They show how a shared threshold produces different retained ranks across matrices, without implying that ranks are proportional to complete-model storage or that a gate-value distribution has been measured.

\end{document}